\documentstyle[aclap]{article}

\newcommand{\smrm}[1]{\mbox{\protect\scriptsize #1}}
\newcommand{\epsfscaledbox}[2]{\psfig{figure=#1,width=#2}}
\def\bo1{\mbox{BO-}1}
\def\boo1{\mbox{BO-o}1}
\def\MLE1{\mbox{MLE-}1}
\def\MLEo1{\mbox{MLE-o}1}
\def\rand{\mbox{RAND}}
\def\conf{P_{\smrm{C}}}
\def\confarg{P_{\smrm{C}}(w_1' | w_1)}
\def\p{p(w_2)}
\def\q{p'(w_2)}

\include{psfig}

\title{\vspace{-75pt}
{\normalsize \tt \hfill Appears in the proceedings of ACL-EACL '97} \\ \mbox{}\\Similarity-Based Methods For Word Sense Disambiguation}
\author{
Ido Dagan \\
Dept. of Mathematics and \\ Computer Science \\
Bar Ilan University \\
 Ramat Gan 52900, Israel\\
{\tt dagan@macs.biu.ac.il} \And
Lillian Lee \\
Div. of Engineering and \\ Applied Sciences \\ Harvard
University \\ Cambridge, MA 01238, USA\\
{\tt llee@eecs.harvard.edu}\And
Fernando Pereira \\
AT\&T Labs -- Research \\ 600 Mountain
Ave. \\ Murray Hill, NJ 07974, USA \\{\tt pereira@research.att.com}}

\begin{document}
\bibliographystyle{fullname}
\maketitle
\begin{abstract}
We compare four similarity-based estimation methods against
back-off and maximum-likelihood estimation methods on a pseudo-word sense
disambiguation task in which we controlled for both unigram and bigram
frequency.  The similarity-based methods perform up to
40\% better on this particular task.  We also conclude that events that 
occur only once in the training set have major impact on
similarity-based estimates.
\end{abstract}

\newtheorem{fact}{Fact}
\newtheorem{definition}{Definition}

\section{Introduction}

The problem of data sparseness affects all statistical methods for
natural language processing.  Even large training sets tend to
misrepresent low-probability events, since rare events may not appear
in the training corpus at all.

We concentrate here on the problem of estimating the probability of
{\em unseen} word pairs, that is, pairs that do not occur in the
training set.  Katz's back-off scheme \cite{Katz:87a}, widely used in
bigram language modeling, estimates the probability of an unseen
bigram by utilizing unigram estimates.  This has the undesirable
result of assigning unseen bigrams the same probability if they are
made up of unigrams of the same frequency.

Class-based methods \cite{Brown:92c,Pereira:93a,Resnik:92a} cluster
words into classes of similar words, so that one can base the estimate of a
word pair's probability on the averaged cooccurrence probability of
the classes to which the two words belong.  However, a word is
therefore modeled by the average behavior of many words, which may
cause the given word's idiosyncrasies to be ignored.  For instance,
the word ``red'' might well act like a generic color word in most
cases, but it has distinctive cooccurrence patterns with respect to
words like ``apple,'' ``banana,'' and so on.

We therefore consider similarity-based estimation schemes that do not
require building general word classes.  Instead, estimates for the
most similar words to a word $w$ are combined; the evidence provided
by word $w'$ is weighted by a function of its similarity to $w$.
Dagan, Markus, and Markovitch (1993) propose such a scheme for
predicting which unseen cooccurrences are more likely than others.
However, their scheme does not assign probabilities.  In what follows,
we focus on probabilistic similarity-based estimation methods.

We compared several such methods, including that of Dag\-an,
Per\-eira, and Lee \shortcite{Dagan:94a} and the {\it cooccurrence
smoothing method} of Essen and Steinbiss \shortcite{Essen:92a},
against classical estimation methods, including that of Katz, in a
decision task involving unseen pairs of direct objects and verbs,
where unigram frequency was eliminated from being a factor.  We found
that all the similarity-based schemes performed almost 40\% better
than back-off, which is expected to yield about 50\% accuracy in our
experimental setting.  Furthermore, a scheme based on the total
divergence of empirical distributions to their average\footnote{To the
best of our knowledge, this is the first use of this particular
distribution dissimilarity function in statistical language
processing.  The function itself is implicit in earlier work on
distributional clustering \cite{Pereira:93a}, has been used by Tishby
(p.c.)  in other distributional similarity work, and, as suggested by
Yoav Freund (p.c.), it is related to results of \newcite{Hoeffding} on
the probability that a given sample was drawn from a given joint
distribution.} yielded statistically significant improvement in error
rate over cooccurrence smoothing.

We also investigated the effect of removing extremely low-frequency
events from the training set.  We found that, in contrast to back-off
smoothing, where such events are often discarded from training with
little discernible effect, similarity-based smoothing methods suffer
noticeable performance degradation when singletons (events that occur
exactly once) are omitted.

\section{Distributional Similarity Models}

We wish to model conditional probability distributions arising from 
the coocurrence of linguistic objects, typically words, in certain 
configurations. We thus consider pairs $(w_{1},w_{2})\in V_{1}\times 
V_{2}$ for appropriate sets $V_{1}$ and $V_{2}$, not necessarily 
disjoint. In what follows, we use subscript $i$ for 
the $i^{th}$ element of a pair; thus $P(w_{2}|w_{1})$ is the conditional 
probability (or rather, some empirical estimate, the true probability 
being unknown) that a pair has second element $w_{2}$ given that its 
first element is $w_{1}$; and $P(w_1 | w_2)$ denotes the
probability estimate, according to the base language model, that $w_1$
is the first word of a pair given that the second word is $w_2$.
$P(w)$ denotes the base estimate for the unigram probability of word
$w$.

A similarity-based language model consists of three parts: a scheme
for deciding which word pairs require a similarity-based estimate, a
method for combining information from similar words, and, of course, a
function measuring the similarity between words.  We give the details
of each of these three parts in the following three sections.  We will
only be concerned with similarity between words in $V_1$.

\subsection{Discounting and Redistribution}
\label{sec:redistribute}

Data sparseness makes the {\em maximum
likelihood estimate (MLE)} for word pair probabilities unreliable.
The MLE for the probability of a word pair $(w_1,w_2)$, conditional on
the appearance of word $w_1$,  is simply
\begin{equation}
P_{ML}(w_2|w_1) = \frac{c(w_1,w_2)}{c(w_1)},
\end{equation}
where $c(w_1,w_2)$ is the frequency of $(w_1,w_2)$ in the training
corpus and $c(w_1)$ is the frequency of $w_1$. However, $P_{ML}$ is
zero for any unseen word pair, which leads to
extremely inaccurate estimates for word pair probabilities.

Previous proposals for remedying the above problem
\cite{Good:53a,Jelinek:92a,Katz:87a,Church+Gale:91a} adjust the MLE in
so that the total probability of seen word pairs is less than one,
leaving some probability mass to be redistributed among the unseen
pairs. In general, the adjustment involves either {\em interpolation},
in which the MLE is used in linear combination with an estimator
guaranteed to be nonzero for unseen word pairs, or {\em discounting},
in which a reduced MLE is used for seen word pairs, with the
probability mass left over from this reduction used to model unseen
pairs.

The discounting approach is
the one adopted by \newcite{Katz:87a}:
\begin{equation}
\hat{P}(w_2| w_1) = \left\{\!\!\!\!
\begin{array}{l@{\hspace{0.6ex}}l}
P_d(w_2 | w_1) & \mbox{$c(w_1,w_2) > 0$} \\
\alpha(w_1)P_r(w_2 | w_1) & \mbox{o.w.}
\end{array}\right.,\label{genmodel}
\end{equation}

\noindent where $P_d$ represents the Good-Turing discounted estimate
\cite{Katz:87a} for seen word pairs, and $P_r$ denotes the model for
probability redistribution among the unseen word pairs.  $\alpha(w_1)$
is a normalization factor.

Following Dagan, Pereira, and Lee \shortcite{Dagan:94a}, we modify Katz's formulation by
writing $P_r(w_2|w_1)$ instead 
$P(w_2)$, enabling us to use similarity-based estimates for unseen
word pairs instead of basing the estimate for the pair on 
unigram frequency $P(w_2)$.  Observe that similarity
estimates are used for unseen word pairs only.

We next investigate estimates for $P_r(w_2| w_1)$ derived by averaging
information from words that are distributionally similar to $w_1$.

\subsection{Combining Evidence}
\label{sec:combine}
Similarity-based models assume that if word $w_1'$
is ``similar'' to word $w_1$, then $w_1'$ can yield information about
the probability of unseen word pairs involving $w_1$.  We use a
weighted average of the evidence provided by similar words, where the
weight given to a particular word $w_1'$ depends on its similarity to
$w_1$.

More precisely, let $W(w_1,w_1')$ denote an increasing function of the
similarity between $w_1$ and $w_1'$, and let ${\cal S}(w_1)$
denote the set of words most similar to $w_1$.  Then the
general form of similarity model we consider is a $W$-weighted linear
combination of predictions of similar words:
\begin{equation}
P_{\smrm{SIM}}(w_2|w_1) = \sum_{w_1' \in {\cal S}(w_1)}
\frac{W(w_1,w_1')}{N(w_1)}{P(w_2|w_1')},
\label{sim-formula}
\end{equation}
where $N(w_1) = \sum_{w_1' \in {\cal S}(w_1)}W(w_1,w_1')$ is a
normalization factor.
According to this formula, $w_2$ is more likely to occur
with $w_1$ if it tends to occur with the words that are most similar to $w_1$.

Considerable latitude is allowed in defining the set ${\cal S}(w_1)$,
as is evidenced by previous work that can be put in the above form.
Essen and Steinbiss \shortcite{Essen:92a} and Karov and Edelman \shortcite{Karov:96a} (implicitly) set ${\cal
S}(w_1) = V_1$. However, it may be desirable to restrict ${\cal
S}(w_1)$ in some fashion, especially if $V_1$ is large.  For instance,
\newcite{Dagan:94a} use the closest $k$ or fewer words $w_1'$ such
that the dissimilarity between $w_1$ and $w_1'$ is less than a
threshold value $t$; $k$ and $t$ are tuned experimentally.

Now, we could directly replace $P_r(w_2|w_1)$ in the back-off
equation (\ref{genmodel}) with $P_{\smrm{SIM}}(w_2|w_1)$.
However, other variations are possible, such as interpolating
with the unigram probability $P(w_2)$:
$$P_r(w_2|w_1) = \gamma P(w_2) + (1 - \gamma) P_{\smrm{SIM}}(w_2|w_1),$$ 
where $\gamma$ is determined experimentally
\cite{Dagan:94a}.  This represents, in effect, a linear combination of
the similarity estimate and the back-off estimate: if $\gamma = 1$,
then we have exactly Katz's back-off scheme.  As we focus in this paper on
alternatives for $P_{\smrm{SIM}}$, we will not consider
this approach here; that is, for the rest of this paper, $P_r(w_2|w_1)
= P_{\smrm{SIM}}(w_2|w_1)$.

\subsection{Measures of Similarity}

We now consider several word similarity functions that can be
derived automatically from the statistics of a training corpus, as
opposed to functions derived from
manually-constructed word classes \cite{Resnik:92a}.  All the
similarity functions we describe below depend just on the base language
model $P(\cdot | \cdot)$, not the discounted model $\hat{P}(\cdot |
\cdot)$ from Section \ref{sec:redistribute} above.

\subsubsection{KL divergence}
{\it Kullback-Leibler (KL) divergence} is a
standard infor\-mation-theoretic measure of the dissimilarity between two
probability mass functions \cite{Cover:91a}.  We can apply it to the
conditional distribution $P( \cdot | w_1)$ induced by $w_1$ on words
in $V_2$:
\begin{equation}
D(w_1 \Vert w_1') = \sum_{w_2} P(w_2|w_1) \log \frac{P(w_2|w_1)}{P(w_2|w_1')}.
\label{KL}
\end{equation}

For $D(w_1\Vert w_1')$ to be defined it must be the case that
$P(w_2|w_1') > 0$ whenever $P(w_2|w_1) > 0$. Unfortunately, this will
not in general be the case for MLEs based on samples, so we would need 
smoothed estimates of $P(w_2|w_1')$ that redistribute some
probability mass to zero-frequency events. However, using smoothed
estimates for $P(w_2|w_1)$ as well requires a sum over all
$w_2\in V_2$, which is expensive for the large vocabularies under
consideration. Given the smoothed denominator distribution, we set
\[
W(w_1,w_1')=10^{-\beta D(w_1||w_1')} \qquad ,\]
where $\beta$ is a free parameter.

\subsubsection{Total divergence to the average}
\label{sec:avg}
A related measure is based on the total KL divergence to the average 
of the two distributions:  
\begin{equation}
A(w_1,w_1') = D \left( w_1 \biggl\Vert \frac{w_1 + w_1'}{2} \right) + D \left(
w_1' \biggl\Vert
\frac{w_1 + w_1'}{2} \right),
\label{avg}
\end{equation} 
where $ (w_1 + w_1')/2$ shorthand for the
distribution 
\[
\frac{1}{2}(P(\cdot|w_1) + P(\cdot|w_1')) \]
Since 
$D(\cdot||\cdot) \geq 0$, $A(w_1, w_1') \geq 0$. Furthermore, 
letting $p(w_{2}) = P(w_{2}|w_{1})$, $p'(w_{2}) = P(w_{2}|w_{1}')$ 
and ${\cal C}=\{w_{2}:p(w_{2})>0,p'(w_{2})>0\}$,
it is 
straightforward to show by grouping terms appropriately that

\begin{eqnarray*}
A(w_1,w_1')   =  & \sum_{w_{2}\in {\cal C}} & \Big\{ H(\p + \q)
\\
& & -{}{}H(\p) - H(\q) \Big\} \\
&  + & 2 \log 2, 
\end{eqnarray*}
where $H(x) = -x \log x$.
Therefore,  $A(w_1,w_1')$ is bounded, ranging between $0$ and $2
\log 2$, and smoothed estimates are not required because probability 
ratios are not involved.  In addition,  the calculation of $A(w_1,w_1')$
requires summing only over those $w_2$ for which $P(w_2|w_1)$ and
$P(w_2|w_1')$ are both non-zero, which, for sparse data, makes
the computation quite fast. 

As in the KL divergence case, we set $W(w_1,w_1')$ to be $10^{-\beta
A(w_1,w_1')}$.

\subsubsection{$L_1$ norm}

The {\it $L_1$ norm} is defined as
\begin{equation}
L(w_1,w_1') = \sum_{w_2} \left| P(w_2|w_1) - P(w_2|w_1') \right|.
\label{var}
\end{equation}
By grouping terms as before, we can
express $L(w_1,w_1')$ in a form depending only on
the ``common'' $w_2$:
\begin{eqnarray*}
L(w_1, w_1') & = & 2 - \sum_{w_2\in C} \p  - \sum_{w_2\in C} \q \\
& &  + \sum_{w_2\in C} |
\p - \q |.
\end{eqnarray*}
This last form makes it clear that $0 \leq L(w_1, w_1') \leq 2$, with equality
if and only if there are no words $w_2$ such that both $P(w_2| w_1)$ and
$P(w_2|w_1')$ are strictly positive.

Since we require a weighting scheme that is decreasing in $L$, we set
\[ W(w_1,w_1') = (2
- L(w_1,w_1'))^\beta\] with  $\beta$ again free.

\subsubsection{Confusion probability}

\newcite{Essen:92a} introduced {\it confusion
probability} \footnote{Actually, they present two alternative
definitions.  We use their model 2-B, which they found yielded the
best experimental results.}, which estimates the probability that word
$w_1'$ can be substituted for word $w_1$:

\begin{eqnarray*}
P_{\smrm{C}}(w_1' | w_1) &= & W(w_1,w_1') \nonumber \\
& = & \sum_{w_2} \frac{P(w_1
| w_2) P(w_1' | w_2) P(w_2)} {P(w_1)}
\end{eqnarray*}
\noindent Unlike the measures described above, $w_1$ may not
necessarily be the ``closest'' word to itself, that is, there may
exist a word $w_1'$ such that $\confarg  > P_{\smrm{C}}(w_1|w_1)$.

The confusion probability can be computed from empirical estimates 
provided all unigram estimates are nonzero (as we assume throughout). 
In fact, the use of smoothed estimates like those of Katz's back-off 
scheme is problematic, because those estimates typically do not 
preserve consistency with respect to marginal estimates and Bayes's 
rule. However, using consistent estimates (such as the MLE), we can 
rewrite $\conf$ as follows:
\begin{displaymath}
\confarg = \sum_{w_2} \frac{P(w_2|w_1)}{P(w_2)} \cdot
P(w_2| w_1') P(w_1').
\end{displaymath}

\noindent This form reveals another important difference between the
confusion probability and the functions $D$, $A$, and $L$ described in
the previous sections.  Those functions rate $w_1'$ as similar to
$w_1$ if, roughly, $P(w_2|w_1')$ is high when $P(w_2 | w_1)$ is.
$\confarg$, however, is greater for those $w_1'$ for which $P(w_1',
w_2)$ is large when $P(w_2|w_1)/P(w_2)$ is.  When the ratio
$P(w_2|w_1)/P(w_2)$ is large, we may think of $w_2$ as being
exceptional, since if $w_2$ is infrequent, we do not expect
$P(w_2|w_1)$ to be large.

\subsubsection{Summary}

Several features of the measures of similarity listed above are
summarized in table \ref{table:simsum}. ``Base LM constraints'' are
conditions that must be satisfied by the probability estimates of the
base language model.  The last column indicates whether the weight
$W(w_1, w_1')$ associated with each similarity function depends on a
parameter that needs to be tuned experimentally.

\begin{table*}[t]
\begin{center}
\begin{tabular}{l|l|c|c}
name &  range &  base LM constraints & tune? \\ \hline
$D$ & $[0, \infty]$ & $P(w_2 | w_1') \neq 0$ if $P(w_2|w_1) \neq 0$ &
yes \\
$A$ & $[0, 2 \log 2]$ &none & yes\\
$L$ & $[0,2]$ & none & yes\\
$P_{\smrm{C}}$ & $[0, \frac{1}{2} \max_{w_2} P(w_2)]$ & Bayes 
consistency & no\\
\end{tabular}
\end{center}
\caption{\label{table:simsum} Summary of similarity function
properties }
\end{table*}

\section{Experimental Results}

We evaluated the similarity measures listed above on a word sense
disambiguation task, in which each method is presented with a noun and
two verbs, and decides which verb is more likely to have the noun as a
direct object.  Thus, we do not measure the absolute quality of the
assignment of probabilities, as would be the case in a perplexity
evaluation, but rather the relative quality.  We are therefore able to
ignore constant factors, and so we neither normalize the similarity
measures nor calculate the denominator in equation
(\ref{sim-formula}).

\subsection{Task: Pseudo-word Sense Disambiguation}

In the usual word sense disambiguation problem, the method to be tested
is presented with an ambiguous word in some context, and is asked to
identify the correct sense of the word from the context.  For example,
a test instance might be the sentence fragment ``robbed the bank'';
the disambiguation method must decide whether ``bank'' refers to a
river bank, a savings bank, or perhaps some other alternative.

While sense disambiguation is clearly an important task, it presents
numerous experimental difficulties.  First, the very notion of
``sense'' is not clearly defined; for instance, dictionaries may
provide sense distinctions that are too fine or too coarse for the
data at hand.  Also, one needs to have training data for which the
correct senses have been assigned, which can require considerable
human effort.  

To circumvent these and other difficulties, we set up a pseudo-word
disambiguation experiment \cite{Schutze92a,Gale92b} the general
format of which is as follows.  We first construct a list of {\em
pseudo-words}, each of which is the combination of two different words
in $V_2$.  Each word in $V_2$ contributes to exactly one pseudo-word.
Then, we replace each $w_2$ in the test set with its corresponding
pseudo-word.  For example, if we choose to create a pseudo-word out of
the words ``make'' and ``take'', we would change the test data like
this:
\begin{center}
\begin{tabular}{llcl}
make & plans & $\Rightarrow$ & \{make, take\} plans \\
take & action &  $\Rightarrow$ & \{make, take\} action \\
\end{tabular}
\end{center}
The method being tested must choose between the two
words that make up the pseudo-word.
\subsection{Data}

We used a statistical part-of-speech tagger \cite{Church:88a} and
pattern matching and concordancing tools (due to David Yarowsky) to
identify transitive main verbs and  head nouns of the corresponding
direct objects in 44 million words of 1988 Associated Press newswire.
We selected the noun-verb pairs for the $1000$ most frequent nouns in
the corpus.  These pairs are undoubtedly somewhat noisy given the
errors inherent in the part-of-speech tagging and pattern matching.

We used $80\%$, or $587833$, of the pairs so derived, for building
base bigram language models, reserving $20\%$ for testing purposes.
As some, but not all, of the similarity measures require smoothed
language models, we calculated both a Katz back-off language model ($P
= \hat{P}$ (equation (\ref{genmodel})), with $P_r(w_2|w_1) = P(w_2)$),
and a maximum-likelihood model ($P = P_{\smrm{ML}}$).  Furthermore, we
wished to investigate Katz's claim that one can delete {\em
singletons}, word pairs that occur only once, from the training set
without affecting model performance \cite{Katz:87a}; our training set
contained $82407$ singletons.  We therefore built four base language
models, summarized in Table \ref{table:langmod}.

\begin{table}[ht]
\begin{center}
\begin{tabular}{c|cc}
 	& with singletons 	& no singletons
 \\ 
 & (587833 pairs)	&(505426 pairs) \\ \hline
MLE	&  \MLE1			& \MLEo1	\\
Katz 	& \bo1			& \boo1			\\
\end{tabular}
\end{center}
\caption{\label{table:langmod} Base Language Models}
\end{table}

Since we wished to test the effectiveness of using similarity for
unseen word cooccurrences, we removed from the test set any
verb-object pairs that occurred in the training set; this resulted in
$17152$ {\em unseen} pairs (some occurred multiple times).  The unseen
pairs were further divided into five equal-sized parts, $T_1$ through
$T_5$, which formed the basis for fivefold cross-validation: in each
of five runs, one of the $T_i$ was used as a performance test set,
with the other 4 sets combined into one set used for tuning parameters
(if necessary) via a simple grid search.  Finally, test pseudo-words
were created from pairs of verbs with similar frequencies, so as to
control for word frequency in the decision task.  We use error rate as
our performance metric, defined as
\[
\frac{1}{N} (\mbox{\# of incorrect choices } + (\mbox{\# of
ties})/2)
\]
where $N$ was the size of the test corpus.  A tie occurs
when the two words making up a pseudo-word are deemed equally likely.

\subsection{Baseline Experiments}

The performances of the four base language models are shown in
table \ref{table:baseline}.  \MLE1 and \MLEo1 both have
error rates of exactly $.5$ because the test sets consist of unseen
bigrams, which are all assigned a probability of $0$ by maximum-likelihood
estimates, and thus are all ties for this method.  The back-off models
\bo1 and \boo1 also perform similarly.

\begin{table}[ht]
\begin{center}
\begin{tabular}{l|lllll}
 & $T_1$ & $T_2$ & $T_3$ & $T_4$ & $T_5$ \\ \hline
\MLE1 & .5 & .5 & .5 & .5 & .5 \\
\MLEo1 & \H{ } & \H{ } & \H{ } & \H{ } & \H{ } \\
\bo1 & 0.517 & 0.520 & 0.512 & 0.513 &0.516 \\
\boo1 & 0.517 & 0.520 & 0.512 & 0.513 &0.516 \\
\end{tabular}
\end{center}
\caption{\label{table:baseline} Base Language Model Error Rates}
\end{table}

Since the back-off models consistently performed worse than the MLE
models, we chose to use only the MLE models in our subsequent
experiments.  Therefore, we only ran comparisons between the measures
that could utilize unsmoothed data, namely, the $L_1$ norm,
$L(w_1,w_1')$; the total divergence to the average, $A(w_1, w_1')$; and the
confusion probability, $P_{\smrm{C}}(w_1' | w_1)$. \footnote{It should
be noted, however, that on \bo1 data, KL-divergence performed slightly
better than the $L_1$ norm.} In the full paper, we give detailed 
examples showing the different neighborhoods induced by the different 
measures, which we omit here for reasons of space.

\subsection{Performance of Similarity-Based Methods}
\label{sec:eval}
\begin{figure}[htb]
\epsfscaledbox{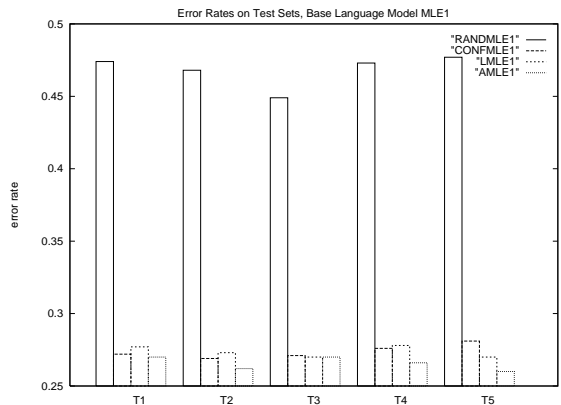}{3in}
\caption{\label{fig:MLE1}  Error rates for each test set, where the
base language model was \MLE1.  The methods, going from left to right,
are $\rand\;$, $\conf$, $L$, and $A$. The performances shown
are for settings of $\beta$ that were optimal for the corresponding
training set.  $\beta$ ranged from $4.0$ to
$4.5$ for $L$ and from $10$ to $13$ for $A$.}
\end{figure}

Figure \ref{fig:MLE1} shows the results on the five test sets, using
\MLE1 as the base language model.  The parameter $\beta$ was always
set to the optimal value for the corresponding training set.  \rand,
which is shown for comparison purposes, simply chooses the weights
$W(w_1, w_1')$ randomly.  ${\cal S}(w_1)$ was set equal to $V_1$ in
all cases.

The similarity-based methods consistently outperform the MLE method
(which, recall, always has an error rate of .5) and Katz's back-off
method (which always had an error rate of about .51) by a huge margin;
therefore, we conclude that information from other word pairs is very
useful for unseen pairs where unigram frequency is not informative.
The similarity-based methods also do much better than $\rand$, which
indicates that it is not enough to simply combine information from
other words arbitrarily: it is quite important to take word similarity
into account.  In all cases, $A$ edged out the other methods.  The
average improvement in using $A$ instead of $\conf$ is .0082; this
difference is significant to the .1 level ($p < .085$), according to
the paired t-test.

\begin{figure}[htb]
\epsfscaledbox{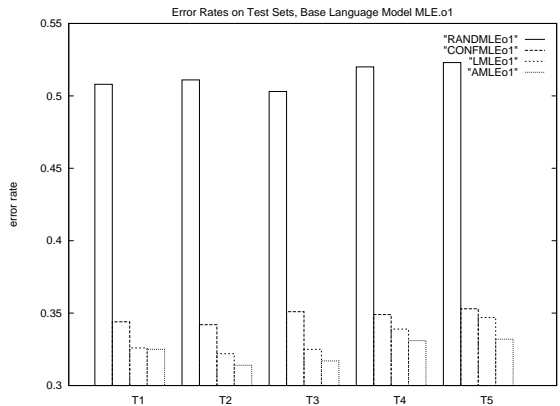}{3in}
\caption{\label{fig:MLEo1}  Error rates for each test set, where the
base language model was \MLEo1.  $\beta$ ranged from $6$ to
$11$ for $L$ and from $21$ to $22$ for $A$.}
\end{figure}

The results for the \MLEo1 case are depicted in figure
\ref{fig:MLEo1}.  Again, we see the similarity-based methods
achieving far lower error rates than the MLE, back-off, and $\rand$
methods, and again, $A$ always performed the best.  However, with
singletons omitted the difference between $A$ and $\conf$ is
even greater, the average difference being $.024$, which is
significant to the .01 level (paired t-test).

An important observation is that all methods, including \rand, were
much more effective if singletons were included in the base language
model; thus, in the case of unseen word pairs, Katz's claim that
singletons can be safely ignored in the back-off model does not hold
for similarity-based models.

\section{Conclusions}

Similarity-based language models provide an
appealing approach for dealing with data sparseness.  We have
described and compared the performance of four such models against two
classical estimation methods, the MLE method and Katz's back-off
scheme, on a pseudo-word disambiguation task.  We observed that the
similarity-based methods perform much better on unseen word pairs,
with the measure based on the KL divergence to the average, being
the best overall.

We also investigated Katz's claim that one can discard singletons in
the training data, resulting in a more compact language model, without
significant loss of performance.  Our results indicate 
that for similarity-based language modeling, singletons are quite
important; their omission leads to significant degradation of performance.

\section*{Acknowledgments}
We thank Hiyan Alshawi, Joshua Goodman, Rebecca Hwa, Stuart Shieber, and Yoram Singer
for many helpful comments and discussions.  Part of this work was done
while the first and second authors were visiting AT\&T Labs.  This
material is based upon work supported in part by the National Science
Foundation under Grant No. IRI-9350192.  The second author also
gratefully acknowledges  support from a National Science Foundation
Graduate Fellowship and an AT\&T GRPW/ALFP grant.

\end {document}